\documentclass[10pt,twocolumn,letterpaper]{article}

\usepackage{cvpr}              

\usepackage{graphicx}
\usepackage{amsmath}
\usepackage{amssymb}
\usepackage{booktabs}
\usepackage{bm}
\usepackage{multicol,multirow}
\usepackage[accsupp]{axessibility}

\usepackage[pagebackref,breaklinks,colorlinks,urlcolor=black]{hyperref}

\usepackage[capitalize]{cleveref}
\crefname{section}{Sec.}{Secs.}
\Crefname{section}{Section}{Sections}
\Crefname{table}{Table}{Tables}
\crefname{table}{Tab.}{Tabs.}

\def\cvprPaperID{1697} 
\def\confName{CVPR}
\def\confYear{2023}

\begin{document}

\title{Q-DETR: An Efficient Low-Bit Quantized Detection Transformer}

\author{Sheng~Xu\textsuperscript{1}$^{\dagger}$, Yanjing~Li\textsuperscript{1}$^{\dagger}$, Mingbao~Lin\textsuperscript{3}, 
Peng~Gao\textsuperscript{4},
Guodong~Guo\textsuperscript{5},
Jinhu~L\"u\textsuperscript{1,2},
Baochang~Zhang\textsuperscript{1,2}$^{*}$\\
\textsuperscript{1} Beihang University \quad
\textsuperscript{2} Zhongguancun Laboratory\quad
\textsuperscript{3} Tencent\\
\textsuperscript{4} Shanghai AI Laboratory \quad 
\textsuperscript{5} UNIUBI Research, Universal Ubiquitous Co. \\
}
\maketitle
\newcommand\blfootnote[1]{%
\begingroup 
\renewcommand\thefootnote{}\footnotetext{#1}%
\addtocounter{footnote}{0}%
\endgroup 
}
\blfootnote{$^{\dagger}$ Equal contribution.}
\blfootnote{$^*$ Corresponding author: bczhang@buaa.edu.cn}
\blfootnote{$^1$ Code: \url{https://github.com/SteveTsui/Q-DETR}}
\begin{abstract}
The recent detection transformer~(DETR) has advanced object detection, but its application on resource-constrained devices requires massive computation and memory resources. Quantization stands out as a solution by representing the network in low-bit parameters and operations. However, there is a significant performance drop when performing low-bit quantized DETR (Q-DETR) with existing quantization methods. We find that the bottlenecks of Q-DETR come from the query information distortion through our empirical analyses. This paper addresses this problem based on a distribution rectification distillation (DRD). We  formulate our DRD as a bi-level optimization problem, which can be derived by generalizing the information bottleneck (IB) principle to the learning of Q-DETR. At the inner level, we conduct a distribution alignment for the queries to maximize the self-information entropy. At the upper level, we  introduce a new foreground-aware query matching scheme to effectively transfer the teacher information to distillation-desired features to minimize the conditional information entropy. Extensive experimental results show that our method performs much better than  prior arts. For example, the 4-bit Q-DETR can theoretically accelerate DETR with ResNet-50 backbone by 6.6$\times$ and achieve 39.4\% AP, with only 2.6\% performance gaps than its real-valued counterpart on the COCO dataset\,$^1$.
\end{abstract}

\section{Introduction}
\label{sec:intro}

Inspired by the success of natural language processing (NLP), object detection with transformers (DETR) has been introduced to train an end-to-end detector via a transformer encoder-decoder~\cite{carion2020end}. Unlike early works~\cite{ren2016faster,liu2016ssd} that often employ convolutional neural networks (CNNs) and require post-processing procedures, {\em e.g.}, non-maximum suppression (NMS), and hand-designed sample selection, DETR treats object detection as a direct set prediction problem.

Despite this attractiveness, DETR usually has a tremendous number of parameters and float-pointing operations (FLOPs). 
For instance, there are 39.8M parameters taking up 159MB memory usage and 86G FLOPs in the DETR model with ResNet-50 backbone~\cite{he2016deep} (DETR-R50). 
This leads to an unacceptable memory and computation consumption during inference, and challenges deployments on devices with limited supplies of resources.


\begin{figure}
    \centering
    \includegraphics[scale=.36]{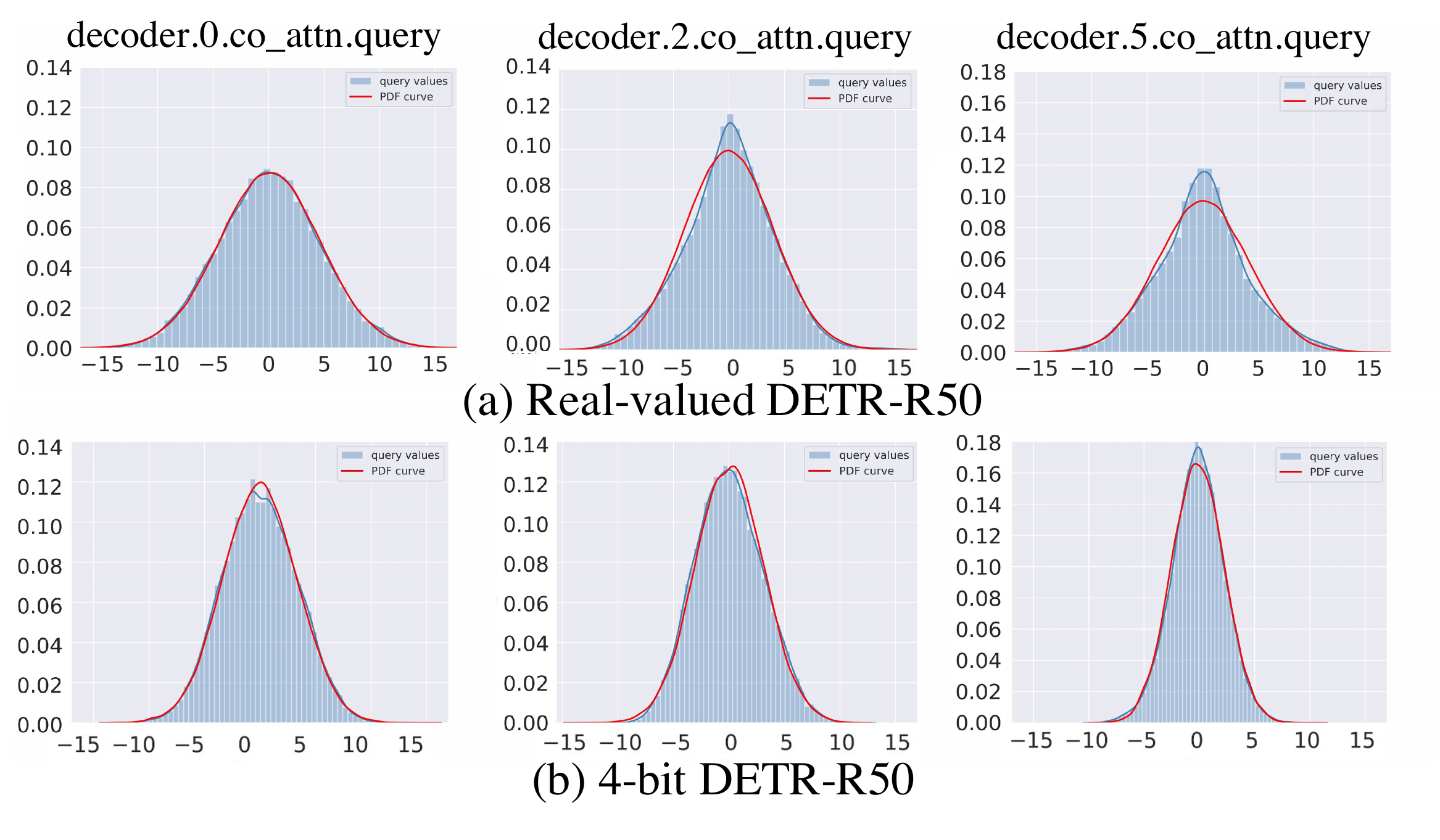} 
    \caption{The histogram of query values ${\bf q}$ (blue shadow) and corresponding PDF curves (red curve) of Gaussian distribution~\cite{li2022q}, {\em w.r.t} the cross attention of different decoder layers in (a) real-valued DETR-R50, and (b) 4-bit quantized DETR-R50 (baseline). 
    Gaussian distribution is generated from the statistical mean and variance of the query values.
    The query in quantized DETR-R50 bears information distortion compared with the real-valued one. Experiments are performed on the VOC dataset~\cite{voc2007}.
    }
    \label{fig:motivation1}
\end{figure}

Therefore, substantial efforts on network compression have been made towards efficient online inference~\cite{denil2013predicting,xu2021layer,xu2022ida,romero2014fitnets}. Quantization is particularly popular for deploying on AI chips by representing a network in low-bit formats.
Yet prior post-training quantization (PTQ) for DETR~\cite{liu2021post} derives quantized parameters from pre-trained real-valued models, which often restricts the model performance in a sub-optimized state due to the lack of fine-tuning on the training data. In particular, the performance drastically drops when quantized to ultra-low bits (4-bits or less).
Alternatively, quantization-aware training (QAT)~\cite{liu2020reactnet,xu2022recurrent} performs quantization and fine-tuning on the training dataset simultaneously, leading to trivial performance degradation even with significantly lower bits. Though QAT methods have been proven to be very effective in compressing CNNs~\cite{liu2018bi,esser2019learned} for computer vision tasks, an exploration of low-bit DETR remains untouched.

\begin{figure}
    \centering
    \includegraphics[scale=.25]{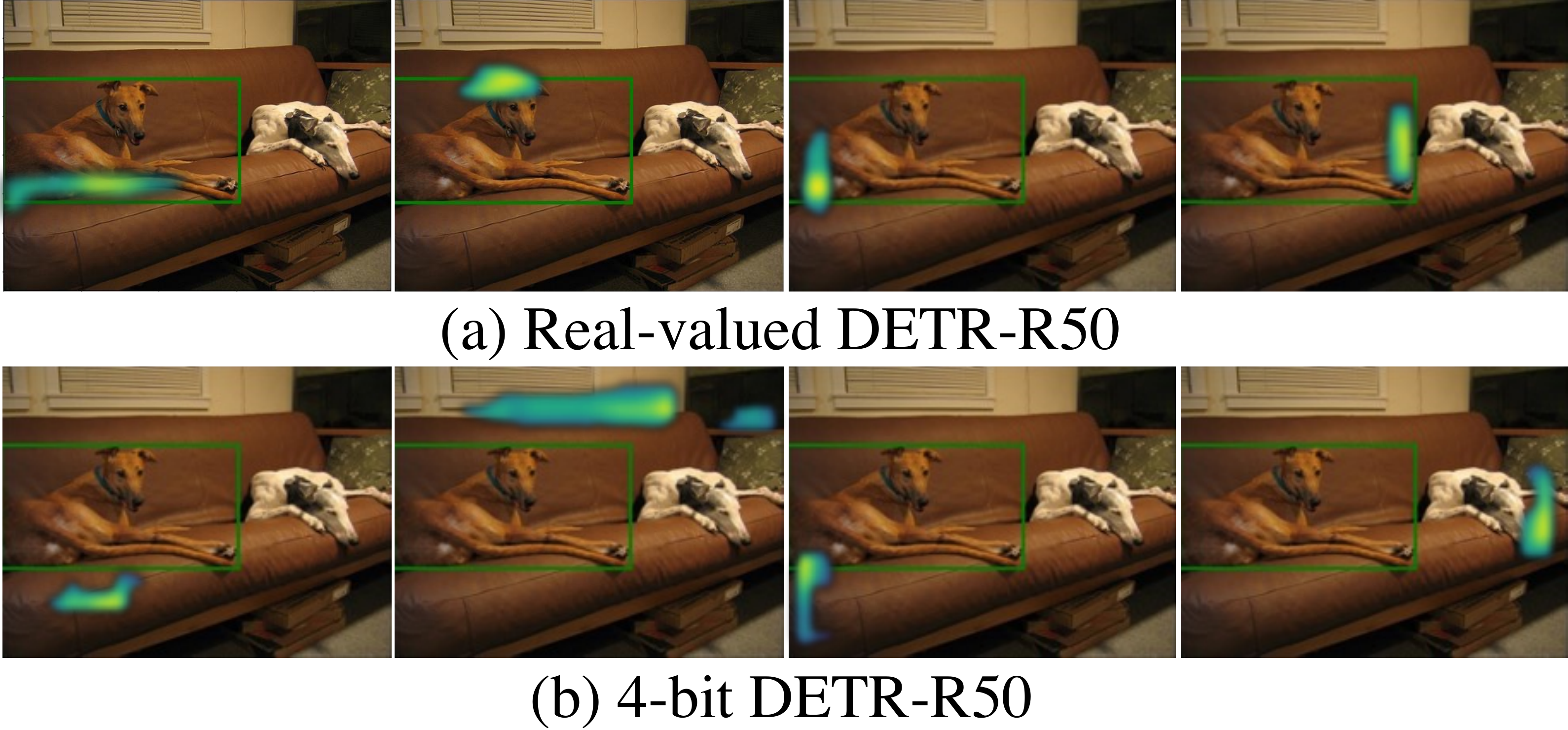} 
    \caption{Spatial attention weight maps in 
    the last decoder of (a) real-valued DETR-R50, and (b) 4-bit quantized DETR-R50.
    The green rectangle denotes the ground-truth bounding box.
    Following~\cite{meng2021conditional}, the highlighted area denotes the large attention weights in the selected four heads in compliance with bound prediction. Compared to its real-valued counterpart that focuses on the ground-truth bounds, quantized DETR-R50 deviates significantly.
    %
    }
    \label{fig:motivation2}
\end{figure}

In this paper, we first build a low-bit DETR baseline, a straightforward solution based on common QAT techniques~\cite{bhalgat2020lsq}. Through an empirical study of this baseline, we observe significant performance drops on the VOC~\cite{voc2007} dataset. For example, a 4-bit quantized DETR-R50 using LSQ~\cite{esser2019learned} only achieves 76.9\% AP$_{50}$, leaving a 6.4\% performance gaps compared with the real-valued DETR-R50. 
We find that the incompatibility of existing QAT methods mainly stems from the unique attention mechanism in DETR, where the spatial dependencies are first constructed between the object queries and encoded features. Then the co-attended object queries are fed into box coordinates and class labels by a feed-forward network. A simple application of existing QAT methods on DETR leads to query information distortion, and therefore the performance severely degrades.
Fig.\,\ref{fig:motivation1} exhibits an example of information distortion in query features of 4-bit DETR-R50, where we can see significant distribution variation of the query modules in quantized DETR and real-valued version.
%
%
%
The query information distortion causes the inaccurate focus of spatial attention, which can be verified by following~\cite{meng2021conditional} to visualize the spatial attention weight maps in 4-bit and real-valued DETR-R50 in Fig.\,\ref{fig:motivation2}. We can see that the quantized DETR-R50 bear's inaccurate object localization. Therefore, a more generic method for DETR quantization is necessary.

\begin{figure*}
\centering
\includegraphics[scale=.28]{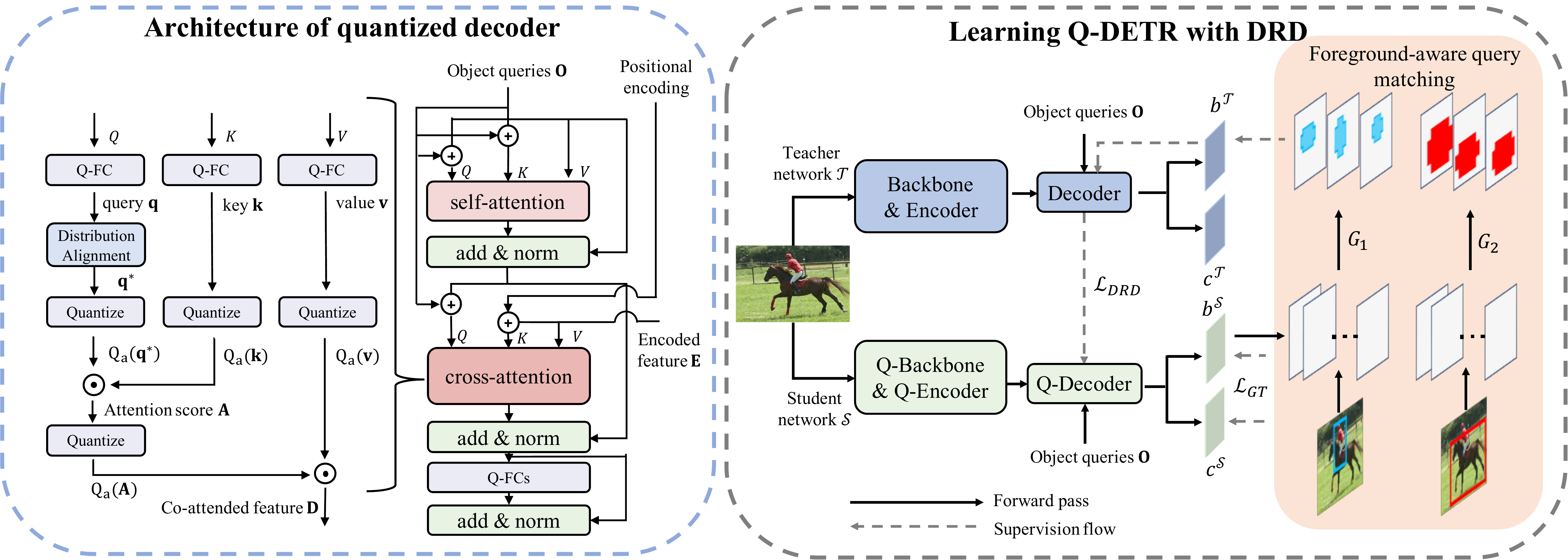} 
\caption{Overview of the proposed Q-DETR framework. We introduce the distribution rectification distillation method (DRD) to refine the performance of Q-DETR. From left to right, we respectively show the detailed decoder architecture of Q-DETR and the learning framework of Q-DETR. The Q-Backbone, Q-Encoder, and Q-Decoder denote quantized architectures, respectively.}
\label{fig:framework}
\end{figure*}

To tackle the issue above, we propose an efficient low-bit quantized DETR (Q-DETR) by rectifying the query information of the quantized DETR as that of the real-valued counterpart. Fig.\,\ref{fig:framework} provides an overview of our Q-DETR, which is mainly accomplished by a distribution rectification knowledge distillation method (DRD). We find ineffective knowledge transferring from the real-valued teacher to the quantized student primarily because of the information gap and distortion. 
Therefore, we formulate our DRD as a bi-level optimization framework established on the information bottleneck principle (IB). Generally, it includes an inner-level optimization to maximize the self-information entropy of student queries and an upper-level optimization to minimize the conditional information entropy between student and teacher queries.
At the inner level, we conduct a distribution alignment for the query guided by its Gaussian-alike distribution, as shown in Fig.\,\ref{fig:motivation1}, leading to an explicit state in compliance with its maximum information entropy in the forward propagation. At the upper level, we introduce a new foreground-aware query matching that filters out low-qualified student queries for exact one-to-one query matching between student and teacher, providing valuable knowledge gradients to push minimum conditional information entropy in the backward propagation.

This paper attempts to introduce a generic method for DETR quantization. The significant contributions in this paper are outlined as follows: 
(1) We develop the first QAT quantization framework for DETR, dubbed Q-DETR.
(2) We use a bi-level optimization distillation framework, abbreviated as DRD.
(3) We observe a significant performance increase compared to existing quantized baselines.

\section{Related Work}

\textbf{Quantization}. Quantized neural networks often possess low-bit (1$\sim$4-bit) weights and activations to accelerate the model inference and save memory. 
For example, DoReFa-Net~\cite{zhou2016dorefa} exploits convolution kernels with low bit-width parameters and gradients to accelerate training and inference. 
TTQ~\cite{zhu2016trained} uses two real-valued scaling coefficients to quantize the weights to ternary values.
Zhuang~\textit{et al.}~\cite{Zhuang_2018_CVPR} present a $2\!\sim\!4$-bit quantization scheme using a two-stage approach to alternately quantize the weights and activations, providing an optimal tradeoff among memory, efficiency, and performance. 
In~\cite{jung2019learning}, the quantization intervals are parameterized, and optimal values are obtained by directly minimizing the task loss of the network. 
ZeroQ~\cite{cai2020zeroq} supports uniform and mixed-precision quantization by optimizing for a distilled dataset which is engineered to match the statistics of the batch normalization across different network layers. 
Xie~\textit{et al.}~\cite{xie2020deep} introduced transfer learning into network quantization to obtain an accurate low-precision model by utilizing Kullback-Leibler (KL) divergence.
Fang~\textit{et al.}~\cite{fang2020post} enabled accurate approximation for tensor values that have bell-shaped distributions with long tails and found the entire range by minimizing the quantization error. 
Li~\textit{et al.}~\cite{li2022q} proposed an information rectification module and distribution-guided distillation to push the bit-width in a quantized vision transformer. At the same time, we address the quantization in DETR from the IB principle.  
The architectural design has also drawn increasing attention using extra shortcut~\cite{liu2018bi}, and parallel parameter-free shortcuts~\cite{liu2020reactnet} for example.

\textbf{Detection Transformer}.
Driven by the success of transformers~\cite{vaswani2017attention}, several researchers have also explored transformer frameworks for vision tasks. The first DETR~\cite{carion2020end} work introduces the Transformer structure based on the attention mechanism for object detection. But the main drawback of DETR lies in the  highly inefficient training process. 
The approachh of another work modifies the multi-head attention mechanism (MHA). Deformable-DETR~\cite{zhu2020deformable} constructs a sparse and point-to-point MHA mechanism using a static point-wise query sampling method around the reference points. SMCA-DETR~\cite{gao2021fast} introduces a Gaussian-distributed spatial function  before formulating a spatially modulated co-attention. DAB-DETR~\cite{liu2022dab} re-defines the query of DETR as dynamic anchor boxes and performs soft ROI pooling layer-by-layer in a cascade manner. DN-DETR~\cite{li2022dn} introduces query denoising into query generation, reducing the bipartite graph matching difficulty and leading to faster convergence. Another set of arts improves DETR methods using additional learning constraints. For example, UP-DETR~\cite{dai2021up} proposes a novel self-supervised loss to enhance the convergence speed and the performance of DETR.

However, prior arts mainly focus on the training efficiency of DETR, few of which have discussed the quantization of DETR. 
{To this end, we first build a  quantized DETR baseline and then address the query information distortion problem   based on the IB principle. Finally, a new KD method based on a foreground-aware query matching scheme is achieved  to solve Q-DETR effectively.
}

\section{The Challenge of Quantizing DETR}

\subsection{Quantized DETR baseline}
We first construct a baseline to study the low-bit DETR since no relevant work has been previously proposed.
To this end, we follow LSQ+~\cite{bhalgat2020lsq} to introduce a general framework of asymmetric activation quantization and symmetric weight quantization:
\begin{equation}\label{eq:quantization}
\small
    \begin{aligned}
    \bm{x}_q = &\lfloor \operatorname{clip}\{\frac{(\bm{x} - z)}{\alpha_x}, Q_n^x, Q_p^x\} \rceil, {\bf w}_{q} = \lfloor \operatorname{clip}\{\frac{{\bf w}}{\alpha_{\bf w}}, Q_n^{\bf w}, Q_p^{\bf w}\} \rceil,
    \\&
    Q_a(x) = \alpha_x \circ \bm{x}_q + z, \;\;\;\;\;\;\;\;\;\;\;\;\, Q_w(x) = \alpha_{\bf w}\circ {\bf w}_{q},
    \end{aligned}
\end{equation}
where $\operatorname{clip}\{y, r_1, r_2\}$ clips the input $y$ with value bounds $r_1$ and $r_2$; the $\lfloor y \rceil$ rounds $y$ to its nearest integer; the $\circ$ denotes the channel-wise multiplication. 
And $Q_n^x = - 2^{a-1}, Q_p^x = 2^{a-1}-1$, $Q_n^{\bf w} = - 2^{b-1}, Q_p^{\bf w} = 2^{b-1}-1$ are the discrete bounds for $a$-bit activations and $b$-bit weights. $x$ generally denotes the activation in this paper, including the input feature map of convolution and fully-connected layers and input of multi-head attention modules. Based on this, we first give the quantized fully-connected layer as: 
\begin{equation}
\small
     \operatorname{Q-FC}(\bm{x}) = Q_a({\bm{x}}) \cdot Q_w({\bf w}) = \alpha_x\alpha_{\bf w} \circ (\bm{x}_q \odot {\bf w}_q + z / \alpha_x \circ {\bf w}_q), 
     \label{q-fc}
\end{equation}
where $\cdot$ denotes the matrix multiplication and $\odot$ denotes the matrix multiplication with efficient bit-wise operations. The straight-through estimator (STE)~\cite{bengio2013estimating} is used to retain the derivation of the gradient in backward propagation.

In DETR~\cite{carion2020end}, the visual features generated by the backbone are augmented with position embedding and fed into the transformer encoder.
%
%
Given an encoder output ${\bf E}$, DETR performs co-attention between object queries ${\bf O}$ and the visual features ${\bf E}$, which are formulated as:
\begin{equation}
    \begin{aligned}
    {\bf q} &= \operatorname{Q-FC}({\bf O}),\;\;{\bf k},{\bf v} = \operatorname{Q-FC}({\bf E})\\
    {\bf A}_i &= \operatorname{softmax}(Q_a({\bf q})_i \cdot Q_a({\bf k})_i^{\top}/\sqrt{d}), \\
    {\bf D}_i &= Q_a({\bf A})_i\cdot Q_a({\bf v})_i,
    \end{aligned}
    \label{decoder}
\end{equation}
where ${\bf D}$ is the multi-head co-attention module, {\em i.e.}, the co-attended feature for the object query. The $d$ denotes the feature dimension in each head.
%
More FC layers transform the decoder's output features of each object query for the final output. Given box and class predictions, the Hungarian algorithm~\cite{carion2020end} is applied between predictions and ground-truth box annotations to identify the learning targets of each object query. 

\begin{figure}[t]
\centering
\includegraphics[scale=.26]{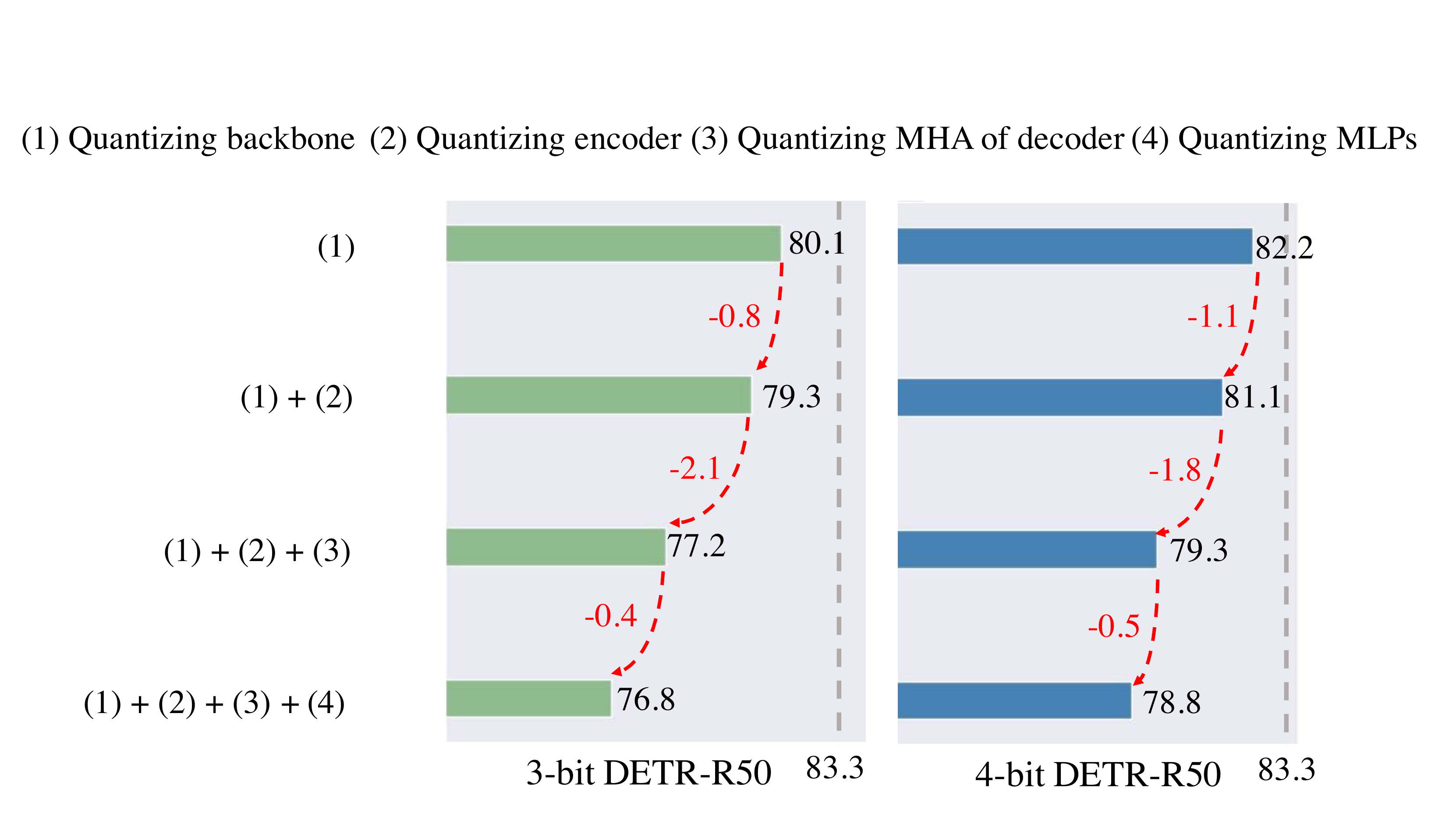}
%
\caption{Performance of 3/4-bit quantized DETR-R50 on VOC with different quantized modules.}
\label{fig:quantization}
\end{figure}

\subsection{Challenge Analysis}
Intuitively, the performance of the quantized DETR baseline largely depends on the information representation capability mainly reflected by the information in the multi-head attention module. 
Unfortunately, such information is severely degraded by the quantized weights and inputs in the forward pass. Also, the rounded and discrete quantization significantly affect the optimization during backpropagation.

We conduct the quantitively ablative experiments by progressively replacing each module of the real-valued DETR baseline with a quantized one and compare the average precision (AP) drop on the VOC dataset~\cite{voc2007} as shown in Fig.\,\ref{fig:quantization}. 
We find that quantizing the MHA decoder module to low bits, {\em i.e.}, (1)+(2)+(3), brings the most significant accuracy drops of accuracy among all parts of the DETR methods, up to 2.1\% in the 3-bit DETR-R50. At the same time, other parts of DETR show comparative robustness to the quantization function. 
Consequently, the critical problem of improving the quantized DETR methods is restoring the information in MHA modules after quantization. Other qualitative results in Fig.\,\ref{fig:motivation1} and Fig.\,\ref{fig:motivation2} also indicate that the degraded information representation is the main obstacle to a better quantized DETR.

\section{The Proposed Q-DETR}

\subsection{Information Bottleneck of Q-DETR}

To address the information distortion of the quantized DETR, we aim to improve the representation capacity of the quantized networks in a knowledge distillation framework. Generally, we utilize a real-valued DETR as a teacher and a quantized DETR as a student, which are distinguished with superscripts $\mathcal{T}$ and $\mathcal{S}$, respectively.

Our Q-DETR  pursues the best tradeoff between performance and compression, which is precisely the goal of the information bottleneck (IB) method through quantifying the mutual information that the intermediate layer contains about the input (less is better) and the desired output (more is better)~\cite{shwartz2017opening,tishby2000information}. 
In our case, the intermediate layer comes from the student, while the desired output includes the ground-truth labels as well as the queries of the teacher for distillation. 
Thus, the objective target of our Q-DETR is:
\begin{equation}
    \begin{aligned}
    \mathop{\min}_{\theta^{\mathcal{S}}} I(X; {\bf E}^{\mathcal{S}}) - \beta I( {\bf E}^{\mathcal{S}}, {\bf q}^{\mathcal{S}}; \bm{y}^{GT}) - \gamma I({\bf q}^{\mathcal{S}}; {\bf q}^{\mathcal{T}}), 
    \end{aligned}
    \label{eq:distill_IB}
\end{equation}
where ${\bf q}^{\mathcal{T}}$ and ${\bf q}^{\mathcal{S}}$ represent the queries in the teacher and student DETR methods as predefined in Eq.\,(\ref{decoder}); $\beta$ and $\gamma$ are the Lagrange multipliers~\cite{shwartz2017opening}; $\theta^{\mathcal{S}}$ is the parameters of the student; and $I(\cdot)$ returns the mutual information of two input variables. 
The first item $I(X; {\bf E}^{\mathcal{S}})$ minimizes information between input and visual features ${\bf E}^{\mathcal{S}}$ to extract task-oriented hints~\cite{wang2020bidet}. The second item $I( {\bf E}^{\mathcal{S}}, {\bf q}^{\mathcal{S}}; \bm{y}^{GT})$  maximizes information between extracted visual features and ground-truth labels for better object detection. These two items can be easily accomplished by common network training and detection loss constraints, such as proposal classification and coordinate regression.

The core issue of this paper is to solve the third item $I({\bf q}^{\mathcal{S}}; {\bf q}^{\mathcal{T}})$, which attempts to address the information distortion in student query via introducing teacher query as a priori knowledge. To accomplish our goal, we first expand the third item and reformulate it as:
\begin{equation}
\begin{aligned}
I({\bf q}^{\mathcal{S}}; {\bf q}^{\mathcal{T}}) = H({\bf q}^{\mathcal{S}}) - H({\bf q}^{\mathcal{S}}|{\bf q}^{\mathcal{T}}), 
\end{aligned}
\label{eq:mutual}
\end{equation}
where $H({\bf q}^{\mathcal{S}})$ returns the self information entropy expected to be maximized while $H({\bf q}^{\mathcal{S}}|{\bf q}^{\mathcal{T}})$ is the conditional entropy expected to be minimized. 
It is challenging to optimize the above maximum \& minimum items simultaneously. Instead, we make a compromise to reformulate Eq.\,(\ref{eq:mutual}) as a bi-level issue~\cite{liu2021investigating,colson2007overview} that alternately optimizes the two items, which is explicitly defined as:
\begin{equation}
\begin{aligned}
& \mathop{\min}_{\theta}  H({\bf q}^{\mathcal{S}^*}|{\bf q}^{\mathcal{T}}), \\
\operatorname{s.t.} \;\;\;\; &{\bf q}^{\mathcal{S}^*} = \mathop{\arg \max}_{{\bf q}^{\mathcal{S}}} H({\bf q}^{\mathcal{S}}). 
\end{aligned}
\label{eq:bi-level}
\end{equation}

Such an objective involves two sub-problems, including an inner-level optimization to derive the current optimal query ${\bf q}^{\mathcal{S}^*}$ and an upper-level optimization to conduct knowledge transfer from the teacher to the student. Below, we show that the two sub-problems can be solved in the forward \& backward network propagation's.


%
%
%

%
%

\subsection{Distribution Rectification Distillation}
\label{sec:DRD}

\textbf{Inner-level optimization}. 
We first detail the maximization of self-information entropy.
According to the definition of self information entropy, $H({\bf q}^{\mathcal{S}})$ can be implicitly expanded as:
\begin{equation}
    H({\bf q}^{\mathcal{S}}) = -\int_{{\bf q}^{\mathcal{S}}_i \in {\bf q}^{\mathcal{S}}} p({\bf q}^{\mathcal{S}}_i){\operatorname{log} p({\bf q}^{\mathcal{S}}_i)}.
\end{equation}

However, an explicit form of $H({\bf q}^{\mathcal{S}})$ can only be parameterized with a regular distribution $p({\bf q}^{\mathcal{S}}_i)$.
Luckily, the statistical results in Fig.\,\ref{fig:motivation1} shows that the query distribution tends to follow a Gaussian distribution, which is also observed in~\cite{li2022q}. This enables us to solve the inner-level optimization in a distribution alignment fashion. To this end, we first calculate the mean $\mu({\bf q}^{\mathcal{S}})$ and variance $\sigma({\bf q}^{\mathcal{S}})$ of query ${\bf q}^{\mathcal{S}}$ whose distribution is then modeled as ${\bf q}^{\mathcal{S}} \sim \mathcal{N}(\mu({\bf q}^{\mathcal{S}}), \sigma({\bf q}^{\mathcal{S}}))$. Then, the self-information entropy of the student query can be proceeded as:%
\begin{equation}
    \begin{aligned}
    H({\bf q}^{\mathcal{S}}) &= -\mathbb{E}[\operatorname{log} \mathcal{N}(\mu({\bf q}^{\mathcal{S}}), \sigma({\bf q}^{\mathcal{S}}))]\\
    &= -\mathbb{E}[\operatorname{log} [{(2\pi{\sigma({{\bf q}^{\mathcal{S}}})}^2)}^{\frac{1}{2}}\operatorname{exp}(-\frac{{({\bf q}^{\mathcal{S}}_i - \mu({\bf q}^{\mathcal{S}}))}^2}{2 {\sigma({{\bf q}^{\mathcal{S}}})^2}})]] \\
    &= \frac{1}{2}\operatorname{log}2\pi{\sigma({{\bf q}^{\mathcal{S}}})}^2.
    \end{aligned}
\end{equation}

The above objective reaches its maximum of $H({\bf q}^{\mathcal{S}^*}) = (1/2) \log 2 \pi e [\sigma({{\bf q}^{\mathcal{S}})}^2 + \epsilon_{{\bf q}^{\mathcal{S}}}]$ when ${\bf q}^{\mathcal{S}^*} = [{\bf q}^{\mathcal{S}} - \mu({\bf q}^{\mathcal{S}})] / [{\sqrt{\sigma{({\bf q}^{\mathcal{S}})}^2 + \epsilon_{{\bf q}^{\mathcal{S}}}}}]$ where $\epsilon_{{\bf q}^{\mathcal{S}}} = 1e^{-5}$ is a small constant added to prevent a zero denominator.
In practice, the mean and variance might be inaccurate due to query data bias. To solve this we use the concepts in batch normalization (BN)~\cite{santurkar2018does,ioffe2015batch} where a learnable shifting parameter $\beta_{{{\bf q}}^{\mathcal{S}}}$ is added to move the mean value. A learnable scaling parameter $\gamma_{{\bf q}^{\mathcal{S}}}$ is multiplied to move the query to the adaptive position. In this situation, we rectify the information entropy of the query in the student as follows:
    \begin{equation}
        \begin{aligned}
        {\bf q}^{\mathcal{S}^*} &= \frac{{\bf q}^{\mathcal{S}} - \mu({\bf q}^{\mathcal{S}})} {{ \sqrt{\sigma{({\bf q}^{\mathcal{S}})}^2 + \epsilon_{{\bf q}^{\mathcal{S}}}}}}\gamma_{{\bf q}^{\mathcal{S}}} + \beta_{{\bf q}^{\mathcal{S}}},
        \end{aligned}
    \label{eq:rectification}
    \end{equation}
    in which case the maximum self-information entropy of student query becomes $H({\bf q}^{\mathcal{S}^*}) = (1/2)\log 2 \pi e [(\sigma^2_{{\bf q}^{\mathcal{S}}} + \epsilon_{{\bf q}^{\mathcal{S}}})/\gamma^2_{{\bf q}^{\mathcal{S}}}]$. 
    %
    Therefore, in the forward propagation, we can obtain the current optimal query ${\bf q}^{\mathcal{S}^*}$ via Eq.\,(\ref{eq:rectification}), after which, the upper-level optimization is further executed as detailed in the following contents.
    

\textbf{Upper-level optimization}.
We continue minimizing the conditional information entropy between the student and the teacher. 
Following DETR~\cite{carion2020end}, we denote the ground-truth labels by $\bm{y}^{GT}=\{c^{GT}_i, b^{GT}_i\}_{i=1}^{N_{gt}}$ as a set of ground-truth objects where $N_{gt}$ is the number of foregrounds, $c_i^{GT}$ and $b_i^{GT}$ respectively represent the class and coordinate (bounding box) for the $i$-th object.
In DETR, each query is associated with an object. Therefore, we can obtain $N$ objects for teacher and student as well, denoted as $\bm{y}^{\mathcal{S}} = \{c^{\mathcal{S}}_j, b^{\mathcal{S}}_j\}_{j=1}^N$ and $\bm{y}^{\mathcal{T}} = \{c^{\mathcal{T}}_j, b^{\mathcal{T}}_j\}_{j=1}^N$.

The minimization of the conditional information entropy requires the student and teacher objects to be in a one-to-one matching. However, it is problematic for DETR due primarily to the sparsity of prediction results and the instability of the query’s predictions~\cite{li2022dn}. We propose a foreground-aware query matching to rectify ``well-matched'' queries to solve this.  Concretely, we match the ground-truth bounding boxes with this student to find the maximum coincidence as:
%
\begin{equation}
    \begin{aligned}
       G_i = \mathop{\max}_{1\leq j \leq N} \operatorname{GIoU}(b^{GT}_{i}, b^{\mathcal{S}}_{j}),
    \end{aligned}
    \label{eq:sigma}
\end{equation}
where $\operatorname{GIoU}(\cdot)$ is the generalized intersection over union function~\cite{rezatofighi2019generalized}. Each $G_i$ reflects the ``closeness'' of student proposals to the $i$-th ground-truth object.
Then, we retain highly qualified student proposals around at least one ground truth to benefit object recognition~\cite{wang2019distilling} as:
\begin{align}
\begin{split}
\small
b_j^{\mathcal{S}} = \left \{
\begin{array}{ll}
    {b}_j^{\mathcal{S}},         \!\!\!\! & {\operatorname{GIoU}}(b^{GT}_{i}, b^{\mathcal{S}}_{j}) > \tau G_i, \,\,\forall\; i  \\
    \varnothing,                                 & \text{otherwise},
\end{array}
\right.
\end{split}
\end{align}
where $\tau$ is a threshold controlling the proportion of distilled queries. After removing object-empty ($\varnothing$) queries in $\tilde{\bm q}^{\mathcal{S}}$, we form a distillation-desired query set of students denoted as $\tilde{{\bm q}}^{\mathcal{S}}$ associated with its object set $\tilde{{\bm y}}^{\mathcal{S}} =  \{\tilde{c}^{\mathcal{S}}_j, \tilde{b}^{\mathcal{S}}_j\}_{j=1}^{\tilde{N}}$. Correspondingly, we can obtain a teacher query set $\tilde{{\bm y}}^{\mathcal{T}} =  \{\tilde{c}^{\mathcal{T}}_j, \tilde{b}^{\mathcal{T}}_j\}_{j=1}^{\tilde{N}}$. For the $j$-th student query, its corresponding teacher query is matched as:
\begin{equation}
\tilde{c}^{\mathcal{T}}_j, \tilde{b}^{\mathcal{T}}_j = \mathop{\arg \max}_{\tilde{c}^{\mathcal{T}}_k, \tilde{b}^{\mathcal{T}}_k}\sum^{N}_{k=1} \mu_1 \operatorname{GIoU}(\tilde{b}^{\mathcal{S}}_{j}, b^{\mathcal{T}}_{k})-\mu_2\|\tilde{b}^{\mathcal{S}}_{j} - b^{\mathcal{T}}_{k}\|_1, 
\end{equation}
where $\mu_1=2$ and $\mu_2=5$ control the matching function, values of which is to follow~\cite{carion2020end}.

Finally, the upper-level optimization after rectification in Eq.\,(\ref{eq:bi-level}) becomes: 
\begin{equation}
\begin{aligned}
 \mathop{\min}_{\theta} H(\tilde{{\bf q}}^{\mathcal{S}^*}|\tilde{{\bf q}}^{\mathcal{T}}). 
\end{aligned}
\label{eq:L_select2}
\end{equation}

Optimizing Eq.\,(\ref{eq:L_select2}) is challenging. Alternatively, we minimize the norm distance between $\tilde{\bf q}^{\mathcal{S}^*}$ and $\tilde{{\bf q}}^{\mathcal{T}}$, optima of which, \emph{i.e.}, $\tilde{\bf q}^{\mathcal{S}^*} = \tilde{\bf q}^{\mathcal{T}}$, is exactly the same with that in Eq.\,(\ref{eq:L_select2}).
Thus, the final loss for our distribution rectification distillation loss becomes: 
\begin{equation}
\mathcal{L}_{DRD}(\tilde{{\bf q}}^{\mathcal{S}^*}, \tilde{{\bf q}}^{\mathcal{T}}) = \mathbb{E}[\|\tilde{\bf D}^{\mathcal{S}^*} - \tilde{\bf D}^{\mathcal{T}}\|_2],
\label{drd}
\end{equation}
where we use the Euclidean distance of co-attented feature $\tilde{\bf D}$ (see Eq. \ref{decoder}) containing the information query $\tilde{\bf q}$ for optimization.

In backward propagation, the gradient updating drives the student queries toward their teacher hints. Therefore we accomplish our distillation.
The overall training losses for our Q-DETR model are:
\begin{equation}
\small
\mathcal{L} = \mathcal{L}_{GT}(\bm{y}^{GT}, \bm{y}^{\mathcal{S}}) + \lambda \mathcal{L}_{DRD}(\tilde{{\bf q}}^{\mathcal{S}^*}, \tilde{{\bf q}}^{\mathcal{T}}),
\label{final}
\end{equation}
where $L_{GT}$ is the common detection loss for missions such as proposal classification and coordinate regression~\cite{carion2020end}, and $\lambda$ is a tradeoff hyper-parameter.

\section{Experiments}
In this section, we evaluate the performance of the proposed Q-DETR mode using popular DETR~\cite{carion2020end} and SMCA-DETR~\cite{gao2021fast} models. To the best of our knowledge, there is no publicly available source code  on quantization-aware training of DETR methods at this point, so we implement the baseline and LSQ~\cite{esser2019learned} methods ourselves.

\subsection{Datasets and Implementation Details}
\label{sec:setup}
\textbf{Datasets}.
We first conduct the ablative study and hyper-parameter selection on the PASCAL VOC dataset \cite{voc2007}, which contains natural images from 20 different classes. We use the VOC {\tt trainval2012}, and VOC {\tt trainval2007} sets to train our model, which contains approximately 16k images, and the VOC {\tt test2007} set to evaluate our Q-DETR, which contains 4952 images. We report COCO-style metrics for the VOC dataset: AP, AP$_{50}$ (default VOC metric), and AP$_{75}$.
We further conduct the experiments on the COCO {\tt 2017} \cite{coco2014} object detection tracking. Specifically, we train the models on COCO {\tt train2017} and evaluate the models on COCO {\tt val2017}.
We list the average precision (AP) for IoUs$\in [0.5:0.05:0.95]$, designated as AP, using COCO's standard evaluation metric. For further analyzing our method, we also list AP$_{50}$, AP$_{75}$, AP$_s$, AP$_m$, and AP$_l$.

\begin{figure}
        \centering
        \begin{subfigure}{0.495\linewidth}
    	\centering
    	\includegraphics[width=1.0\linewidth]{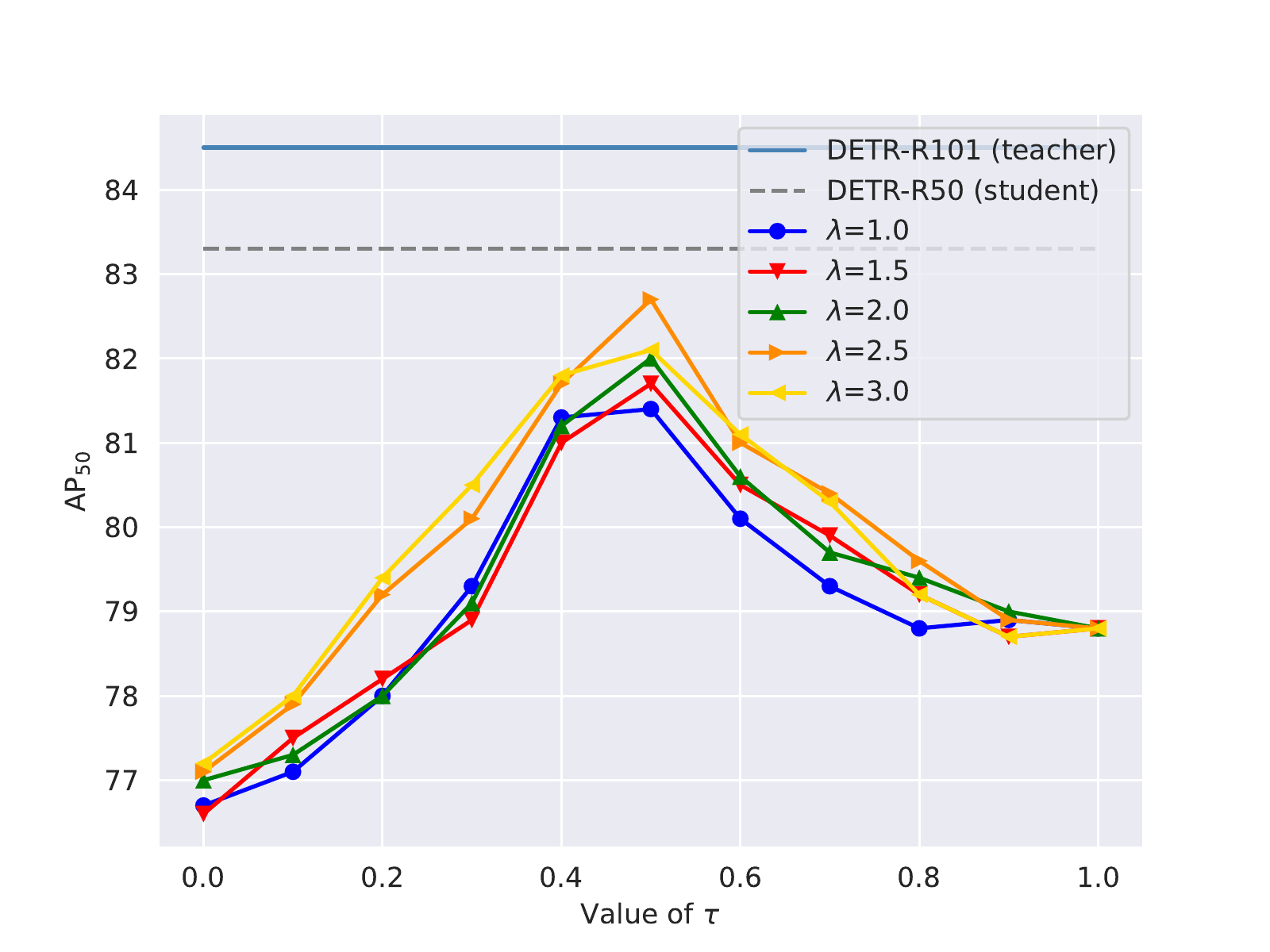}
    	\caption{Effect of $\tau$ and $\lambda$.}
    	\label{motivation:teacher1}
        \end{subfigure}
        \begin{subfigure}{0.495\linewidth}
    	\centering
    	\includegraphics[width=1.0\linewidth]{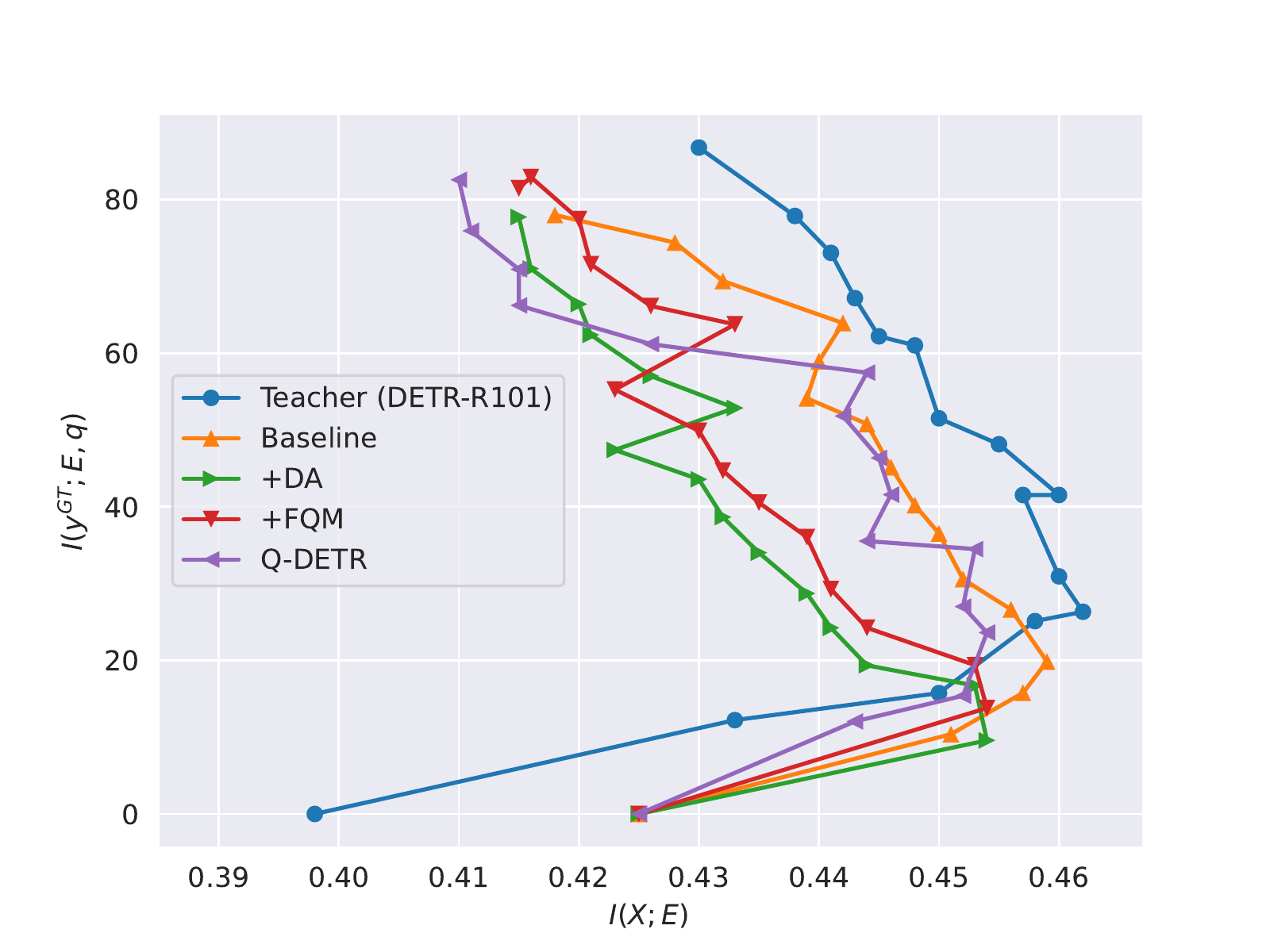}
 		\caption{Mutual information curves.}
    	\label{motivation:teacher_false1}
        \end{subfigure}
        \caption{(a) We select $\tau$ and $\lambda$ using 4-bit Q-DETR-R50 on VOC. (b) The mutual information curves of $I(X; {\bf E})$ and $I(\bm{y}^{GT}; {\bf E}, {\bf q})$ (Eq.\;\ref{eq:distill_IB}) on the information plane. The red curves represent the teacher model (DETR-R101). The orange, green, red, and purple lines represent the 4-bit baseline, 4-bit baseline + DA, 4-bit baseline + FQM, and 4-bit baseline + DA + FQM (4-bit Q-DETR).}
        \label{hyper-parameter}
    \end{figure}

\textbf{Implementation Details}. 
Our Q-DETR is trained with the DETR~\cite{carion2020end} and SMCA-DETR~\cite{gao2021fast} framework. We select the ResNet-50 \cite{he2016deep} and modify it with Pre-Activation structures and RPReLU \cite{liu2020reactnet} function following \cite{liu2022nonuniform}. PyTorch \cite{paszke2017automatic} is used for implementing Q-DETR. We run the experiments on 8 NVIDIA Tesla A100 GPUs with $80$ GB memory. We use ImageNet ILSVRC12 \cite{imagenet12} to pre-train the backbone of a quantized student. The training protocol is the same as the employed frameworks \cite{carion2020end, gao2021fast}. 
Specifically, we use a batch size of 16. AdamW \cite{loshchilov2017decoupled} is used to optimize the Q-DETR, with the initial learning rate of $1e^{-4}$.
We train for 300/500 epochs for the Q-DETR on VOC/COCO dataset, and the learning rate is multiplied by 0.1 at the 200/400-th epoch, respectively. Following the SMCA-DETR, we train the Q-SMCA-DETR for 50 epochs, and the learning rate is multiplied by 0.1 at the 40-th epoch on both the VOC and COCO datasets. We utilize a multi-distillation strategy, where we save the encoder and decoder network as real-valued at the first stage. Then we train the fully quantized DETR at the second stage, where we load the weight from the checkpoint of first stage. 
We select real-valued DETR-R101 (84.5\% AP$_{50}$ on VOC and 43.5\% AP on COCO) and SMCA-DETR-R101 (85.3\% AP$_{50}$ on VOC and 44.4\% AP on COCO) as teacher network.

\subsection{Ablation Study}

\textbf{Hyper-parameter selection}.  
As mentioned above, we select hyper-parameters $\tau$ and $\lambda$ in this part using the 4-bit Q-DETR model. We show the model performance (AP$_{50}$) with different setups of hyper-parameters $\{\tau,\lambda\}$ in Fig.\,\ref{hyper-parameter} (a), where we conduct ablative experiments on the baseline + DA (AP$_{50}$=78.8\%).
As can be seen, the performances increase first and then decrease with the increase of $\tau$ from left to right. Since $\tau$ controls the proportion of selected distillation-desired queries, we show that the full-imitation ($\tau=0$) performs worse than the vanilla baseline with no distillation ($\tau=1$), showing query selection is necessary. The figure also shows that the performances increase first and then decrease with the increase of $\tau$ from left to right. The Q-DETR obtains better performances with $\tau$ set as 0.5 and 0.6. 
With the varying value of $\lambda$, we find $\{\lambda,\tau\}$ = \{2.5, 0.6\} boost the performance of Q-DETR most, achieving 82.7\% AP on VOC {\tt test2007}. Based on the ablative study above, we set hyper-parameters $\tau$ and $\lambda$ as 0.6 and 2.5 for the experiments in this paper.

\textbf{Effectiveness of components}. We show quantitative improvements of components in Q-DETR in Tab.\,\ref{tab:ablation}.
As shown in Tab.\,\ref{tab:ablation}, the quantized DETR baseline suffers a severe performance drop on AP$_{50}$ (13.6\%, 6.5\%, and 5.3\% with 2/3/4-bit, respectively). DA and FQM improve the performance when used alone, and the two techniques further boost the performance considerably when combined. For example, the DA improves the 2-bit baseline by 1.9\%, and the FQM achieves a 5.2\% performance improvement. While combining the DA and FQM, the performance improvement achieves 6.7\%. 

\textbf{Information analysis}. We further show the information plane following \cite{wang2021revisiting} in Fig.\;\ref{hyper-parameter}(b). We adopt the test AP$_{50}$ to quantify $I(\bm{y}^{GT}; {\bf E}, {\bf q})$. We employ a reconstruction decoder to decode the encoded feature ${\bf E}$ to reconstruct the input and quantify $I(X;{\bf E})$ using the $\ell_1$ loss. As shown in  Fig.\;\ref{hyper-parameter}(b), the curve of the larger teacher DETR-R101 is usually on the right of the curve of small student models, which indicates a greater ability of information representation. Likewise, the purple line (Q-DETR-R50) is usually on the right of the three left curves, showing the information representation improvements with the proposed methods.


\begin{table}[]
\centering
\caption{Evaluating the components of Q-DETR-R50 on the VOC dataset. \#Bits (W-A-Attention) denotes the bit-width of weights, activations, and attention activations. DA denotes the distribution alignment module. FQM denotes foreground-aware query matching.}
\setlength{\tabcolsep}{1.1mm}{
\begin{tabular}{ccccccc}
\hline
Method                                                              & \#Bits   & AP$_{50}$     & \#Bits & AP$_{50}$     & \#Bits & AP$_{50}$     \\ \hline
Real-valued                                                         & 32-32-32 & 83.3          & -      & -             & -      & -             \\ \hline
Baseline                                                            & 4-4-8    & 78.0          & 3-3-8  & 76.8          & 2-2-8  & 69.7          \\
+DA                                                                 & 4-4-8    & 78.8          & 3-3-8  & 78.0          & 2-2-8  & 71.6          \\
+FQM                                                                & 4-4-8    & 81.5          & 3-3-8  & 80.9          & 2-2-8  & 74.9          \\
\textbf{\begin{tabular}[c]{@{}c@{}}+DA+FQM\\ (Q-DETR)\end{tabular}} & 4-4-8    & \textbf{82.7} & 3-3-8  & \textbf{82.1} & 2-2-8  & \textbf{76.4} \\ \hline
\end{tabular}}
\label{tab:ablation}
\end{table}

\begin{table}[]
\centering
\caption{We report AP, AP$_{50}$, and AP$_{75}$ ($\%$) with state-of-the-art quantization methods on DETR and SMCA-DETR using VOC {\tt test2007}. \#Bits (W-A-Attention) denotes the bit-width of weights, activations, and attention activations.}
\setlength{\tabcolsep}{1.0mm}{
\begin{tabular}{cccccc}
\hline
Model                                                                      & Method          & \#Bits                 & AP            & AP$_{50}$     & AP$_{75}$     \\ \hline
\multirow{12}{*}{DETR-R50}                                                 & Real-valued     & 32-32-32               & 59.5          & 83.3          & 64.7          \\ \cline{2-6} 
& Percentile      & \multirow{2}{*}{8-8-8} & 54.7          & 79.2          & 60.1          \\
& VT-PTQ          &                        & 57.6          & 82.3          & 63.1          \\ \cline{2-6} 
& LSQ             & \multirow{3}{*}{4-4-8} & 49.7          & 76.9          & 53.0          \\
& Baseline        &                        & 51.3          & 78.0          & 54.1          \\
& \textbf{Q-DETR} &                        & \textbf{57.1} & \textbf{82.7} & \textbf{61.5} \\ \cline{2-6} 
& LSQ             & \multirow{3}{*}{3-3-8} & 47.0          & 75.3          & 49.1          \\
& Baseline        &                        & 49.2          & 76.8          & 51.8          \\
& \textbf{Q-DETR} &                        & \textbf{56.8} & \textbf{82.1} & \textbf{61.2} \\ \cline{2-6} 
& LSQ             & \multirow{3}{*}{2-2-8} & 42.6          & 68.2          & 44.8          \\
& Baseline        &                        & 44.0          & 69.7          & 45.8          \\
& \textbf{Q-DETR} &                        & \textbf{50.7} & \textbf{76.4} & \textbf{54.1} \\ \hline
\multirow{12}{*}{\begin{tabular}[c]{@{}c@{}}SMCA-DETR\\ -R50\end{tabular}} & Real-valued     & 32-32-32               & 56.7          & 83.7          & 62.0          \\ \cline{2-6} 
& Percentile      & \multirow{2}{*}{8-8-8} & 54.7          & 79.2          & 60.1          \\
& VT-PTQ          &                        & 55.9          & 83.0          & 61.3          \\ \cline{2-6} 
& LSQ             & \multirow{3}{*}{4-4-8} & 49.6          & 78.6          & 53.4          \\
& Baseline        &                        & 50.7          & 79.5          & 55.4          \\
& \textbf{Q-DETR}          &                        & \textbf{56.2} & \textbf{83.3} & \textbf{61.6} \\ \cline{2-6} 
& LSQ             & \multirow{3}{*}{3-3-8} &      47.7         &    76.5           &       51.7        \\
& Baseline        &                        &     49.9          &      77.5         &      53.6         \\
& \textbf{Q-DETR}          &                        &       \textbf{54.3}       & \textbf{82.6}     & \textbf{59.5}     \\ \cline{2-6} 
& LSQ             & \multirow{3}{*}{2-2-8} &        42.3       &      69.7         &      44.8         \\
& Baseline        &                        &     43.9          &    70.4           &      46.1         \\
& \textbf{Q-DETR}          &                        &        \textbf{50.2}       &        \textbf{76.7}       &        \textbf{52.6}       \\ \hline
\end{tabular}}
\label{voc}
\end{table}

\subsection{Results on PASCAL VOC}

We first compare our method with the 2/3/4-bit baseline and  LSQ~\cite{esser2019learned} based on the same frameworks for object detection task with the VOC dataset. We also report the detection performance of the 8-bit post-training quantization networks, such as percentile \cite{lin2021fq}, VT-PTQ~\cite{liu2021post}. We use the input resolution following \cite{carion2020end}, {\em i.e.} 1333$\times$800. We mainly discuss the AP$_{50}$ (default VOC metric) in the following. 

We evaluate the proposed Q-DETR on DETR-R50 models in Tab.\,\ref{voc}. For the DETR-R50 model, compared with the 8-bit PTQ method, our 4-bit Q-DETR achieves a much larger compression ratio than 8-bit VT-PTQ, but with a bit of performance improvement (82.7\% {\em vs.} 82.3\%). Also, the proposed method boosts the performance of 2/3/4-bit baseline by 6.7\%, 5.3\%, and 4.7\% with the same architecture and bit-width, which significantly validates the effectiveness of our method.

\begin{table*}[!t]
\centering
\caption{Comparison with state-of-the-art quantization methods using DETR and SMCA-DETR on COCO {\tt val2017}. \#Bits (W-A-Attention) denotes bit-width of weights, activations, and attention activations.}
\begin{tabular}{ccccccccccc}
\hline
Model                           & Method          & \#Bits                 & Size$_{\rm (MB)}$        & OPs$_{\rm (G)}$          & AP & AP$_{50}$     & AP$_{75}$     & AP$_{s}$      & AP$_{m}$      & AP$_{l}$      \\ \hline
\multirow{12}{*}{DETR-R50}      & Real-valued     & 32-32-32               & 159.32                 & 85.51                  & 42.0                                                       & 62.4          & 44.2          & 20.5          & 45.8          & 61.1          \\ \cline{2-11} 
& Percentile      & \multirow{2}{*}{8-8-8} & \multirow{2}{*}{39.83} & \multirow{2}{*}{23.01} & 38.6                                                       & -             & -             & -             & -             & -             \\
& VT-PTQ          &                        &                        &                        & 41.2                                                       & -             & -             & -             & -             & -             \\ \cline{2-11} 
& LSQ             & \multirow{3}{*}{4-4-8} & \multirow{3}{*}{19.92} & \multirow{3}{*}{13.02} & 33.3                                                       & 53.7          & 33.9          & 12.8          & 37.0          & 51.6          \\
& Baseline        &                        &                        &                        & 34.1                                                       & 55.3          & 35.4          & 14.3          & 38.0          & 53.8          \\
& \textbf{Q-DETR} &                        &                        &                        & \textbf{39.4}                                              & \textbf{60.2} & \textbf{41.4} & \textbf{17.7} & \textbf{43.4} & \textbf{59.9} \\ \cline{2-11} 
& LSQ             & \multirow{3}{*}{3-3-8} & \multirow{3}{*}{15.03}  & \multirow{3}{*}{7.61}  & 31.0                                                       & 52.3          & 32.1          & 11.3          & 33.9          & 48.5          \\
& Baseline        &                        &                        &                        & 32.3                                                       & 52.2          & 32.9          & 12.3          & 35.4          & 50.3          \\
& \textbf{Q-DETR} &                        &                        &                        & \textbf{36.1}                                              & \textbf{55.9} & \textbf{37.5} & \textbf{14.6} & \textbf{39.4} & \textbf{55.2} \\ \cline{2-11} 
& LSQ             & \multirow{3}{*}{2-2-8} & \multirow{3}{*}{10.03}  & \multirow{3}{*}{5.32}  & 24.7                                                       & 44.6          & 26.5          & 6.3           & 25.3          & 42.7          \\
& Baseline        &                        &                        &                        & 26.6                                                       & 46.6          & 26.5          & 8.4           & 28.2          & 44.4          \\
& \textbf{Q-DETR} &                        &                        &                        & \textbf{31.4}                                              & \textbf{51.3} & \textbf{31.6} & \textbf{11.6} & \textbf{34.3} & \textbf{49.6} \\ \hline
\multirow{12}{*}{SMCA-DETR-R50} & Real-valued     & 32-32-32               & 164.75                 & 86.65                  & 41.0                                                       & 62.2          & 43.6          & 21.9          & 44.3          & 59.1          \\ \cline{2-11} 
& Percentile      & \multirow{2}{*}{8-8-8} & \multirow{2}{*}{41.19} & \multirow{2}{*}{23.66} & 37.5                                                       & 58.5          & 40.1          & 17.6          & 39.1          & 55.9          \\
& VT-PTQ          &                        &                        &                        & 40.2                                                       & 61.0          & 42.6          & 20.3          & 42.9          & 57.7          \\ \cline{2-11} 
& LSQ             & \multirow{3}{*}{4-4-8} & \multirow{3}{*}{20.59} & \multirow{3}{*}{13.48} & 33.9                                                       & 55.0          & 35.0          & 13.2          & 37.2          & 51.4          \\
& Baseline        &                        &                        &                        & 35.0                                                       & 56.4          & 36.4          & 15.6          & 38.3          & 52.5          \\
& \textbf{Q-DETR} &                        &                        &                        & \textbf{38.3}                                              & \textbf{59.7} & \textbf{39.8} & \textbf{17.7} & \textbf{41.7} & \textbf{56.8} \\ \cline{2-11} 
& LSQ             & \multirow{3}{*}{3-3-8} & \multirow{3}{*}{15.68} & \multirow{3}{*}{8.05}  & 30.1                                                       & 52.6          & 31.4          & 11.9          & 33.4          & 46.6          \\
& Baseline        &                        &                        &                        & 31.8                                                       & 53.7          & 32.6          & 12.6          & 35.2          & 49.8          \\
& \textbf{Q-DETR} &                        &                        &                        & \textbf{35.0}                                              & \textbf{56.3}          & \textbf{36.9}          & \textbf{15.0} & \textbf{39.0} & \textbf{53.1} \\ \cline{2-11} 
& LSQ             & \multirow{3}{*}{2-2-8} & \multirow{3}{*}{10.84}  & \multirow{3}{*}{4.54}  & 23.9                                                       & 42.2          & 24.2          & 9.4           & 26.2          & 37.5          \\
& Baseline        &                        &                        &                        & 25.4                                                       & 44.3          & 25.2          & 8.4           & 27.2          & 40.3          \\
& \textbf{Q-DETR} &                        &                        &                        & \textbf{30.5}                                              & \textbf{51.8} & \textbf{31.8} & \textbf{12.0} & \textbf{33.2} & \textbf{48.0} \\ \hline
\end{tabular}
\label{COCO}
\end{table*}

Besides, our method generates convincing results on SMCA-DETR. As shown in Tab.\,\ref{voc}, the performance of the proposed Q-DETR with SMCA-DETR-R50 outperforms the 2/3/4-bit Baseline method by 6.3\% , 5.1\% and 3.8\% on AP$_{50}$, a large margin. Compared with 8-bit post-training quantization methods, our method achieves a significantly higher compression rate and comparable performance. 

\subsection{Results on COCO}

We further show comparison on the large-scale COCO \cite{coco2014} dataset. We compare our method with the 2/3/4-bit baseline and  LSQ~\cite{esser2019learned} based on the same frameworks. We also report the detection performance of the 8-bit post-training quantization networks, such as percentile \cite{lin2021fq} , VT-PTQ~\cite{liu2021post}. The AP with different IoU thresholds, and AP of objects with varying scales are all reported in Tab.\,\ref{COCO}. 

Tab.\,\ref{COCO} lists the comparison of several quantization approaches and detection frameworks in computing complexity, storage cost. Our Q-DETR significantly accelerates computation and reduces storage requirements for various detectors. 
We follow~\cite{wang2020bidet} to calculate memory usage, by adding 32$\times$ the number of real-valued weights and $a\times$ the number of quantized weights in the $a$-bit networks. 
The number of operations (OPs) is calculated in the same way as~\cite{wang2020bidet}. Current CPUs can handle both bit-wise XNOR and bit-count operations in parallel. The respective number of FLOPs adds $\{\frac{1}{32},\frac{1}{16},\frac{1}{8}\}$ of the number of $\{$2,3,4$\}$-bit multiplications equals the OPs following \cite{liu2020bi}. 

We summarize the experimental results on COCO {\tt val2017} of Q-DETR-R50 from lines 2 to 17 in Tab.\,\ref{COCO}. For the DETR-R50 model, compared with the 8-bit PTQ method, our 4-bit Q-DETR achieves a much larger acceleration than the 8-bit VT-PTQ but with an acceptable performance gap. Also, the proposed method boosts the performance of 2/3/4-bit baseline by 4.8\%, 3.8\% and 5.1\% AP with the same architecture and bit-width, which is significant on the large-scale COCO dataset. Compared with the real-valued counterparts, the proposed 2/3/4-bit Q-DETR achieves computation acceleration and storage savings by 16.07$\times$/11.23$\times$/6.57$\times$ and 15.88$\times$/10.60$\times$/7.99$\times$. The above results are of great significance in the real-time inference of object detection. All of the improvements have impacts on object detection.

For the SMCA-DETR-R50 model, we observe similar performance improvements and compression ratios. For example, the 4-bit Q-SMCA-DETR-R50 theoretically accelerates 6.42$\times$ with only a 2.7\% performance gap compared with the real-valued counterpart, which is significant for real-time DETR methods.

\section{Conclusion}
This paper introduces a novel method for training quantized DETR (Q-DETR) with knowledge distillation to rectify the query distribution. Q-DETR generalizes the information bottleneck (IB) principle and leads a bi-level distribution rectification distillation. We effectively employ a distribution alignment module to solve inner-level and a foreground-aware query matching scheme to solve upper level. As a result, Q-DETR significantly boosts performance of low-bit DETR. Extensive experiments show that Q-DETR surpasses state-of-the-arts in DETR quantization.

\section{Acknowledgements}
This work was supported by National Natural Science Foundation of China under Grant 62141604, 62076016, Beijing Natural Science Foundation L223024.

{\small
\bibliographystyle{ieee_fullname}
\bibliography{egbib}

\begin{thebibliography}{10}\itemsep=-1pt

\bibitem{bengio2013estimating}
Yoshua Bengio, Nicholas L{\'e}onard, and Aaron Courville.
\newblock Estimating or propagating gradients through stochastic neurons for
  conditional computation.
\newblock {\em arXiv preprint arXiv:1308.3432}, 2013.

\bibitem{bhalgat2020lsq}
Yash Bhalgat, Jinwon Lee, Markus Nagel, Tijmen Blankevoort, and Nojun Kwak.
\newblock Lsq+: Improving low-bit quantization through learnable offsets and
  better initialization.
\newblock In {\em Proc. of CVPR Workshops}, pages 696--697, 2020.

\bibitem{cai2020zeroq}
Yaohui Cai, Zhewei Yao, Zhen Dong, Amir Gholami, Michael~W Mahoney, and Kurt
  Keutzer.
\newblock Zeroq: A novel zero shot quantization framework.
\newblock In {\em Proc. of CVPR}, pages 13169--13178, 2020.

\bibitem{carion2020end}
Nicolas Carion, Francisco Massa, Gabriel Synnaeve, Nicolas Usunier, Alexander
  Kirillov, and Sergey Zagoruyko.
\newblock End-to-end object detection with transformers.
\newblock In {\em Proc. of ECCV}, pages 213--229, 2020.

\bibitem{colson2007overview}
Beno{\^\i}t Colson, Patrice Marcotte, and Gilles Savard.
\newblock An overview of bilevel optimization.
\newblock {\em Annals of operations research}, 153(1):235--256, 2007.

\bibitem{dai2021up}
Zhigang Dai, Bolun Cai, Yugeng Lin, and Junying Chen.
\newblock Up-detr: Unsupervised pre-training for object detection with
  transformers.
\newblock In {\em Proc. of CVPR}, pages 1601--1610, 2021.

\bibitem{denil2013predicting}
Misha Denil, Babak Shakibi, Laurent Dinh, Marc'Aurelio Ranzato, and Nando
  De~Freitas.
\newblock Predicting parameters in deep learning.
\newblock In {\em Proc. of NeurIPS}, pages 2148--2156, 2013.

\bibitem{esser2019learned}
Steven~K Esser, Jeffrey~L McKinstry, Deepika Bablani, Rathinakumar Appuswamy,
  and Dharmendra~S Modha.
\newblock Learned step size quantization.
\newblock {\em arXiv preprint arXiv:1902.08153}, 2019.

\bibitem{voc2007}
Mark Everingham, Luc Van~Gool, Christopher~KI Williams, John Winn, and Andrew
  Zisserman.
\newblock The pascal visual object classes (voc) challenge.
\newblock {\em International Journal of Computer Vision}, 88(2):303--338, 2010.

\bibitem{fang2020post}
Jun Fang, Ali Shafiee, Hamzah Abdel-Aziz, David Thorsley, Georgios Georgiadis,
  and Joseph~H Hassoun.
\newblock Post-training piecewise linear quantization for deep neural networks.
\newblock In {\em Proc. of ECCV}, pages 69--86, 2020.

\bibitem{gao2021fast}
Peng Gao, Minghang Zheng, Xiaogang Wang, Jifeng Dai, and Hongsheng Li.
\newblock Fast convergence of detr with spatially modulated co-attention.
\newblock In {\em Proc. of ICCV}, pages 3621--3630, 2021.

\bibitem{he2016deep}
Kaiming He, Xiangyu Zhang, Shaoqing Ren, and Jian Sun.
\newblock Deep residual learning for image recognition.
\newblock In {\em Proc. of CVPR}, pages 770--778, 2016.

\bibitem{ioffe2015batch}
Sergey Ioffe and Christian Szegedy.
\newblock Batch normalization: Accelerating deep network training by reducing
  internal covariate shift.
\newblock In {\em Proc. of ICML}, pages 448--456, 2015.

\bibitem{jung2019learning}
Sangil Jung, Changyong Son, Seohyung Lee, Jinwoo Son, Jae-Joon Han, Youngjun
  Kwak, Sung~Ju Hwang, and Changkyu Choi.
\newblock Learning to quantize deep networks by optimizing quantization
  intervals with task loss.
\newblock In {\em Proc. of CVPR}, pages 4350--4359, 2019.

\bibitem{imagenet12}
Alex Krizhevsky, Ilya Sutskever, and Geoffrey~E Hinton.
\newblock Imagenet classification with deep convolutional neural networks.
\newblock In {\em Proc. of NeurIPS}, pages 1097--1105, 2012.

\bibitem{li2022dn}
Feng Li, Hao Zhang, Shilong Liu, Jian Guo, Lionel~M Ni, and Lei Zhang.
\newblock Dn-detr: Accelerate detr training by introducing query denoising.
\newblock In {\em Proc. of CVPR}, pages 13619--13627, 2022.

\bibitem{li2022q}
Yanjing Li, Sheng Xu, Baochang Zhang, Xianbin Cao, Peng Gao, and Guodong Guo.
\newblock Q-vit: Accurate and fully quantized low-bit vision transformer.
\newblock In {\em Proc. of NeurIPS}, pages 1--12, 2022.

\bibitem{coco2014}
Tsung-Yi Lin, Michael Maire, Serge Belongie, James Hays, Pietro Perona, Deva
  Ramanan, Piotr Doll{\'a}r, and C~Lawrence Zitnick.
\newblock Microsoft coco: Common objects in context.
\newblock In {\em Proc. of ECCV}, pages 740--755, 2014.

\bibitem{lin2021fq}
Yang Lin, Tianyu Zhang, Peiqin Sun, Zheng Li, and Shuchang Zhou.
\newblock Fq-vit: Post-training quantization for fully quantized vision
  transformer.
\newblock In {\em Proc. of IJCAI}, pages 1173--1179, 2021.

\bibitem{liu2021investigating}
Risheng Liu, Jiaxin Gao, Jin Zhang, Deyu Meng, and Zhouchen Lin.
\newblock Investigating bi-level optimization for learning and vision from a
  unified perspective: A survey and beyond.
\newblock {\em IEEE Transactions on Pattern Analysis and Machine Intelligence},
  2021.

\bibitem{liu2022dab}
Shilong Liu, Feng Li, Hao Zhang, Xiao Yang, Xianbiao Qi, Hang Su, Jun Zhu, and
  Lei Zhang.
\newblock Dab-detr: Dynamic anchor boxes are better queries for detr.
\newblock pages 1--19, 2022.

\bibitem{liu2016ssd}
Wei Liu, Dragomir Anguelov, Dumitru Erhan, Christian Szegedy, Scott Reed,
  Cheng-Yang Fu, and Alexander~C Berg.
\newblock Ssd: Single shot multibox detector.
\newblock In {\em Proc. of ECCV}, pages 21--37, 2016.

\bibitem{liu2022nonuniform}
Zechun Liu, Kwang-Ting Cheng, Dong Huang, Eric~P Xing, and Zhiqiang Shen.
\newblock Nonuniform-to-uniform quantization: Towards accurate quantization via
  generalized straight-through estimation.
\newblock In {\em Proc. of CVPR}, pages 4942--4952, 2022.

\bibitem{liu2020bi}
Zechun Liu, Wenhan Luo, Baoyuan Wu, Xin Yang, Wei Liu, and Kwang-Ting Cheng.
\newblock Bi-real net: Binarizing deep network towards real-network
  performance.
\newblock {\em International Journal of Computer Vision}, 128(1):202--219,
  2020.

\bibitem{liu2020reactnet}
Zechun Liu, Zhiqiang Shen, Marios Savvides, and Kwang-Ting Cheng.
\newblock Reactnet: Towards precise binary neural network with generalized
  activation functions.
\newblock In {\em Proc. of ECCV}, pages 143--159, 2020.

\bibitem{liu2021post}
Zhenhua Liu, Yunhe Wang, Kai Han, Wei Zhang, Siwei Ma, and Wen Gao.
\newblock Post-training quantization for vision transformer.
\newblock {\em Proc. of NeurIPS}, pages 1--12, 2021.

\bibitem{liu2018bi}
Zechun Liu, Baoyuan Wu, Wenhan Luo, Xin Yang, Wei Liu, and Kwang-Ting Cheng.
\newblock Bi-real net: Enhancing the performance of 1-bit cnns with improved
  representational capability and advanced training algorithm.
\newblock In {\em Proc. of ECCV}, pages 722--737, 2018.

\bibitem{loshchilov2017decoupled}
Ilya Loshchilov and Frank Hutter.
\newblock Decoupled weight decay regularization.
\newblock In {\em Proc. of ICLR}, pages 1--18, 2017.

\bibitem{meng2021conditional}
Depu Meng, Xiaokang Chen, Zejia Fan, Gang Zeng, Houqiang Li, Yuhui Yuan, Lei
  Sun, and Jingdong Wang.
\newblock Conditional detr for fast training convergence.
\newblock In {\em Proc. of ICCV}, pages 3651--3660, 2021.

\bibitem{paszke2017automatic}
Adam Paszke, Sam Gross, Soumith Chintala, Gregory Chanan, Edward Yang, Zachary
  DeVito, Zeming Lin, Alban Desmaison, Luca Antiga, and Adam Lerer.
\newblock Automatic differentiation in pytorch.
\newblock In {\em Proc. of NeurIPS Workshops}, pages 1--4, 2017.

\bibitem{ren2016faster}
Shaoqing Ren, Kaiming He, Ross Girshick, and Jian Sun.
\newblock Faster r-cnn: Towards real-time object detection with region proposal
  networks.
\newblock {\em IEEE Transactions on Pattern Analysis and Machine Intelligence},
  39(6):1137--1149, 2016.

\bibitem{rezatofighi2019generalized}
Hamid Rezatofighi, Nathan Tsoi, JunYoung Gwak, Amir Sadeghian, Ian Reid, and
  Silvio Savarese.
\newblock Generalized intersection over union: A metric and a loss for bounding
  box regression.
\newblock In {\em Proc. of CVPR}, pages 658--666, 2019.

\bibitem{romero2014fitnets}
Adriana Romero, Nicolas Ballas, Samira~Ebrahimi Kahou, Antoine Chassang, Carlo
  Gatta, and Yoshua Bengio.
\newblock Fitnets: Hints for thin deep nets.
\newblock In {\em Proc. of ICLR}, pages 1--13, 2015.

\bibitem{santurkar2018does}
Shibani Santurkar, Dimitris Tsipras, Andrew Ilyas, and Aleksander Madry.
\newblock How does batch normalization help optimization?
\newblock In {\em Proc. of NeurIPS}, pages 1--11, 2018.

\bibitem{shwartz2017opening}
Ravid Shwartz-Ziv and Naftali Tishby.
\newblock Opening the black box of deep neural networks via information.
\newblock {\em arXiv:1703.00810}, 2017.

\bibitem{tishby2000information}
Naftali Tishby, Fernando~C Pereira, and William Bialek.
\newblock The information bottleneck method.
\newblock {\em arXiv preprint physics/0004057}, 2000.

\bibitem{vaswani2017attention}
Ashish Vaswani, Noam Shazeer, Niki Parmar, Jakob Uszkoreit, Llion Jones,
  Aidan~N Gomez, {\L}ukasz Kaiser, and Illia Polosukhin.
\newblock Attention is all you need.
\newblock In {\em Proc. of NeurIPS}, pages 1--11, 2017.

\bibitem{wang2019distilling}
Tao Wang, Li Yuan, Xiaopeng Zhang, and Jiashi Feng.
\newblock Distilling object detectors with fine-grained feature imitation.
\newblock In {\em Proc. of CVPR}, pages 4933--4942, 2019.

\bibitem{wang2021revisiting}
Yulin Wang, Zanlin Ni, Shiji Song, Le Yang, and Gao Huang.
\newblock Revisiting locally supervised learning: an alternative to end-to-end
  training.
\newblock In {\em Proc. of ICLR}, pages 1--21, 2021.

\bibitem{wang2020bidet}
Ziwei Wang, Ziyi Wu, Jiwen Lu, and Jie Zhou.
\newblock Bidet: An efficient binarized object detector.
\newblock In {\em Proc. of CVPR}, pages 2049--2058, 2020.

\bibitem{xie2020deep}
Zheng Xie, Zhiquan Wen, Jing Liu, Zhiqiang Liu, Xixian Wu, and Mingkui Tan.
\newblock Deep transferring quantization.
\newblock In {\em Proc. of ECCV}, pages 625--642, 2020.

\bibitem{xu2022recurrent}
Sheng Xu, Yanjing Li, Tiancheng Wang, Teli Ma, Baochang Zhang, Peng Gao, Yu
  Qiao, Jinhu L{\"u}, and Guodong Guo.
\newblock Recurrent bilinear optimization for binary neural networks.
\newblock In {\em Proc. of ECCV}, pages 19--35, 2022.

\bibitem{xu2022ida}
Sheng Xu, Yanjing Li, Bohan Zeng, Teli Ma, Baochang Zhang, Xianbin Cao, Peng
  Gao, and Jinhu L{\"u}.
\newblock Ida-det: An information discrepancy-aware distillation for 1-bit
  detectors.
\newblock In {\em Proc. of ECCV}, pages 346--361, 2022.

\bibitem{xu2021layer}
Sheng Xu, Junhe Zhao, Jinhu Lu, Baochang Zhang, Shumin Han, and David Doermann.
\newblock Layer-wise searching for 1-bit detectors.
\newblock In {\em Proc. of CVPR}, pages 5682--5691, 2021.

\bibitem{zhou2016dorefa}
Shuchang Zhou, Yuxin Wu, Zekun Ni, Xinyu Zhou, He Wen, and Yuheng Zou.
\newblock Dorefa-net: Training low bitwidth convolutional neural networks with
  low bitwidth gradients.
\newblock {\em arXiv preprint arXiv:1606.06160}, 2016.

\bibitem{zhu2016trained}
Chenzhuo Zhu, Song Han, Huizi Mao, and William~J Dally.
\newblock Trained ternary quantization.
\newblock In {\em Proc. of ICLR}, pages 1--10, 2017.

\bibitem{zhu2020deformable}
Xizhou Zhu, Weijie Su, Lewei Lu, Bin Li, Xiaogang Wang, and Jifeng Dai.
\newblock Deformable detr: Deformable transformers for end-to-end object
  detection.
\newblock In {\em Proc. of ICLR}, pages 1--16, 2020.

\bibitem{Zhuang_2018_CVPR}
Bohan Zhuang, Chunhua Shen, Mingkui Tan, Lingqiao Liu, and Ian Reid.
\newblock Towards effective low-bitwidth convolutional neural networks.
\newblock In {\em Proc. of CVPR}, pages 7920--7928, 2018.

\end{thebibliography}
}

\end{document}